\documentclass{article}

\usepackage[final]{neurips_2020}

\usepackage[utf8]{inputenc} %
\usepackage[T1]{fontenc}    %
\usepackage{hyperref}       %
\usepackage{url}            %
\usepackage{booktabs}       %
\usepackage{amsfonts}       %
\usepackage{nicefrac}       %
\usepackage{microtype}      %
\usepackage{amsmath}
\usepackage{enumitem}
\usepackage{amssymb}
\usepackage{multirow}
\usepackage{graphicx}
\usepackage{subfigure}
\usepackage[usenames,dvipsnames]{xcolor}
\usepackage{natbib}
\usepackage{esvect}
\definecolor{shadecolor}{gray}{0.9}

\newcommand{\mc}[1]{\multicolumn{1}{c}{#1}}

\title{User-Dependent Neural Sequence Models for Continuous-Time Event Data
} %

\author{%
  Alex Boyd$^1$\quad\textbf{Robert Bamler}$^2$\quad\textbf{Stephan Mandt}$^{1,2}$\quad\textbf{Padhraic Smyth}$^{1,2}$ \\
  {}$^1$\text{Department of Statistics}\quad{}$^2$\text{Department of Computer Science}\\
  University of California, Irvine\\
  \texttt{\{alexjb, rbamler, mandt\}@uci.edu}\quad \texttt{smyth@ics.uci.edu} \\
}

\begin{document}

\maketitle

\begin{abstract}
Continuous-time event data are common in applications such as individual behavior data, financial transactions, and medical health records. Modeling such data can be very challenging, in particular for applications with many different types of events,  since it requires a model to predict the event types as well as the time of occurrence. 
Recurrent neural networks that parameterize time-varying intensity functions
are the current state-of-the-art for predictive modeling with such data. These models typically assume that all event sequences come from the same data distribution. However, in many applications  event sequences are generated by different sources, or \emph{users}, and their characteristics can be very different. In this paper, we extend the broad class of neural marked point process models to mixtures of latent embeddings, %
where each mixture component models the characteristic traits of a given user. Our approach relies on augmenting these models with a latent variable that encodes user characteristics, represented by a mixture model over user behavior that is trained via amortized variational inference. 
We evaluate our methods on four large real-world datasets and demonstrate systematic improvements from our approach over existing work for a variety of predictive metrics such as log-likelihood, next event ranking, and source-of-sequence identification. 
\end{abstract}

\section{Introduction}

Event sequences in continuous time occur across many contexts, leading to a variety of data analysis applications such as forecasting consumer purchases, fraud detection in transaction data, and prediction in clinical medicine.
In such data, each event is typically associated with one of $K$ event types, also known as \emph{marks}, and a timestamp. 
There has been significant amount of prior work in statistics for clustering \citep{du2015dirichlet, xu2017dirichlet}, factorizing \citep{schein2015bayesian}, and generative modeling of such data, typically under the framework of marked temporal point process (MTPP) models \citep{daley2007introduction}. 
We are primarily interested in the third of these objectives. 
The multivariate Hawkes process \citep{hawkes1971spectra,liniger2009multivariate}, Poisson-network \citep{rajaram2005poisson}, piecewise-continuous conditional intensity model \citep{gunawardana2011model}, and proximal graphic event model \citep{bhattacharjya2018proximal} are some examples of the many different MTPP models previously explored.
MTPP models characterize the instantaneous rate of occurrence of each event type, the so-called \emph{intensity function}, conditioned on the history of past events.
However, strong parametric assumptions in these traditional models limit their flexibility for modeling real-world phenomena.

Recent work in machine learning has sought to address these limitations via the use of deep recurrent neural networks (RNNs). These models, such as \citet{du2016recurrent}, use expressive representations for the intensity function,  use event embeddings to avoid parameter explosion, and optimize the associated log-likelihood via stochastic gradient methods.  %
A variety of approaches have been explored to address the complex mix of discrete events and continuous time that occur in real-world event sequences \citep{mei2017neural,wang2017marked,zhang2019self,turkmen2019fastpoint}. In general, these neural-based MTPPs have been found empirically to provide systematically better predictions than their traditional counterparts, due to their more flexible representations for capturing the influence of past events on future ones (both their time stamps and their types), as well as being better able to handle the significant data sparsity associated with large event vocabularies. 
However, a common implicit assumption in these approaches is that all of the modeled sequences originate from the same source (or user), which is often not the case in real-world datasets such as online consumer data or medical history records.
Sufficiently powerful neural-based MTPPs can internally adjust for this heterogeneity after conditioning on a significant portion of a history; however, they exhibit large predictive uncertainty at the beginning of sequences. Thus, it is important to develop techniques that \emph{personalize} predictions to account for  heterogeneity across users.

To develop personalized MTPPs, we propose using variational autoencoders (VAEs) \citep{KW2014}  in conjunction with previous neural-based MTPPs. VAEs are well-suited to address the problem of personalization (e.g., \cite{liang2018variational}) since they distinguish between global and local parameters, which are treated differently during training/inference.
In our setup, global model parameters capture properties that are common across all sequences regardless of associated user.
By contrast, local parameters describe user-specific traits and preferences. They therefore have to be inferred from fewer, user specific data, motivating a Bayesian treatment by the VAE.
We further employ a mixture-of-experts approach \citep{shi2019variational} to account for heterogeneity within each user. 
We demonstrate that our proposed scheme yields systematic improvements in a variety of tasks on four large, real world datasets.

\section{Personalized Event Sequences}\label{sec:problem_statement}

\paragraph{Problem Statement}
We consider the problem of modeling sequences of events $(t,k)$ that occur at irregular times $t$.
Each event carries a mark $k$ corresponding to one of a finite number $K$ of different possible event types.
Since all our training sets are finite, each event sequence is bounded by some time horizon $T>0$, i.e., it can be written as a finite   history  sequence of the form
\begin{align}
    \mathcal H_T = \big( (t_1,k_1), (t_2,k_2), \ldots, (t_{|\mathcal H_T|}, k_{|\mathcal H_T|}) \big)
\end{align}
with times $0\leq t_1 < t_2 < \cdots < t_{|\mathcal H_T|} \leq T$ and marks $k_i\in\{1,\ldots,K\} \;\forall i$. For brevity, let $\mathcal{H} \equiv \mathcal{H}_T$.

\begin{figure}[t]
\begin{center}
\includegraphics[width=1.0\columnwidth]{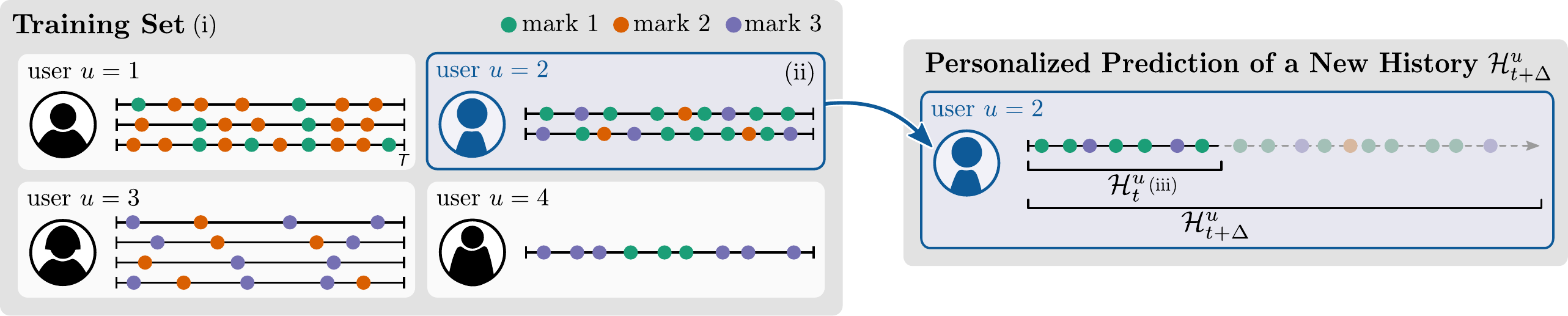}
\end{center}
\caption{Personalized event sequence.
We predict (or evaluate the likelihood of) the semi-transparently drawn sequence on the right conditioned on (i)~a training set (left), (ii)~a few reference sequences from the same user (left, blue), and (iii)~a prefix $\mathcal H_t^u$ of the current sequence (right).}
\label{fig:overview}
\end{figure}

Since our goal is personalization, we assume that each event sequence is associated with a \emph{user}%
\footnote{More generally, ``user'' denotes any source of event sequences, e.g., an individual, organization, or
system.}~$u$.
Our objective is to forecast (or evaluate the likelihood of) an event sequence for a given user~$u$ conditioned on three sources of information, illustrated in Figure~\ref{fig:overview}:
(i)~a large \emph{training set} of event sequences from many users, possibly including~$u$ (Figure~\ref{fig:overview}, left);
(ii)~a smaller (possibly empty) set of \emph{reference sequences} from the same user~$u$, that may or may not be part of the training set (blue box in left part of Figure~\ref{fig:overview}); and
(iii)~a \emph{prefix}, i.e., a (possibly empty) partial sequence $\mathcal H_t^u$ of events performed by user~$u$ just before the time $t$ where we start predicting (solid circles in right part of Figure~\ref{fig:overview}).
Conditioned on these three sources of information, we aim to predict the continuation of the full sequence $\mathcal H_{t+\Delta}^u$ (half-transparent circles in Figure~\ref{fig:overview}, right) for some remaining time span~$\Delta$.

For example, in an online store, the training set (i)~contains many shopping sessions by many users;
the reference sequences (ii)~are previous sessions of a given user~$u$;
and the prefix (iii)~is the already observed part of a currently ongoing session. This setup is quite general, covering many applications with recurring users, e.g., social media platforms,  music streaming services, or e-commerce websites. %

The above enumeration arranges the three types of input information (i)-(iii) as nested subsets organized from global to local context.
It thus reflects the assumption that the predicted event sequence $\mathcal H_{t+\Delta}^u$ follows some general characteristics (e.g., similarities or incompatibilities between different event types) that can be learned from the entire data set~(i).
At the same time, we assume that $\mathcal H_{t+\Delta}^u$ also exhibits some individual traits of the associated user~$u$, which can therefore only be inferred from reference sequences~(ii) from the same user.
Finally, a user's current goal or mood may still vary somewhat over time and can therefore only be inferred from very recent observations in the prefix $\mathcal H_t^u$~(iii).

While the relevance for the current prediction increases as we go from (i) to~(iii), the data size shrinks from (i) to~(iii).
This motivates different treatments of the data sources (i)-(iii) in our models in terms of point estimated global model parameters, a user-specific Bayesian inference variable $z^u$, and the local state of a recurrent neural network, respectively.
The rest of this section provides an overview over the proposed framework, deferring a more detailed discussion to Section~\ref{sec:model}.

\paragraph{Overview of the Proposed Solution}
Forecasting discrete event sequences amounts to both predicting an ordering of future event \emph{types}~$k_i$ as well as their \emph{time stamps}~$t_i$. This is a challenging problem due to the strong correlations between these two data modalities. 
We consider a broad class of stochastic dynamics models---\emph{neural marked temporal point processes} (neural MTPPs)---%
that autoregressively generate events one after another.
For the $i^\text{th}$ event, we draw its time stamp $t_i$ and mark $k_i$ conditioned on the hidden state $h_i$ of a ``decoder'' recurrent neural network (RNN). %
The RNN state~$h_i$ gets updated after each generated event, conditioned on the event it just generated, and on an embedding vector $z^u$ that represents the user~$u$.
This leads to the following stochastic recursion:
\begin{equation}
\begin{gathered}
\label{eq:stoch_rnn}
    (t_{i+1},k_{i+1}) \sim p_\theta(t_{i+1}, k_{i+1} \,|\, h_i, z^u)
    \\\text{with}\quad
    h_{i} = f_\text{Dec}(h_{i-1}, [\mathbf{k}_i;z^u], t_{i}; \eta) 
    \quad \text{and}\quad h_0=\tanh(W_0z^u + b_0)
\end{gathered}
\end{equation}

where $[\,\cdot\,;\cdot\,]$ denotes vector concatenation, $\mathbf{k}_i$ is a learnable continuous vector embedding for the mark~$k_i$ of the $i^\text{th}$ event, 
and $f_\text{Dec}$ is the recurrent unit of the decoder neural network $\textsc{Dec}_\theta$ with learnable parameters~$\theta$ that include $\eta, W_0,$ and $b_0$. %

Apart from the user embedding $z^u$, which is specific to our personalization scheme, Eq.~\ref{eq:stoch_rnn} covers several MTPP models in the literature.
Section~\ref{sec:model} summarizes models that fit into our framework. Note that we are not considering any side-information from users (e.g., age, location, gender, etc.); however, depending on the application these should be straightforward to incorporate into the framework.

Eq.~\ref{eq:stoch_rnn} models the generation of event sequences probabilistically: 
each successive event $(t_{i+1},k_{i+1})$ is sampled from~$p_\theta(t_{i+1},k_{i+1} \,|\, h_i,z^u)$ rather than being generated deterministically.
Probabilistic models allow for the ability to estimate complex statistics, such as the expected time until next event of a specific type, how much more likely event A will come before event B, etc.
Many of these questions would be hard or impossible to answer with a deterministic model.

Eq.~\ref{eq:stoch_rnn} further reflects our problem setting with diverse information sources (i)-(iii) discussed above.
The decoder parameters~$\theta$ are identical for all generated sequences and can thus be trained on the entire data set~(i).
The user embedding~$z^u$ stays constant within each event sequence but varies from user to user.
Thus, $z^u$ has to be inferred from reference sequences~(ii) from the same user. 
By concatenating the user embedding to the mark embedding, we effectively create personalized representations of events for that user. Additionally, by computing the initial stochastic state $h_0$ from $z^u$, we allow for personalized predictions across the entire time window from $t=0$ to $t=T$.
Finally, when completing a partial sequence, the prefix~(iii) can be encoded into the initial RNN state~$h_i$ by unrolling the RNN update (second line of Eq.~\ref{eq:stoch_rnn}) on the events in the prefix.

\section{Model Parameterization and Inference}
\label{sec:model}

\begin{figure}
    \centering
    \includegraphics[width=\linewidth]{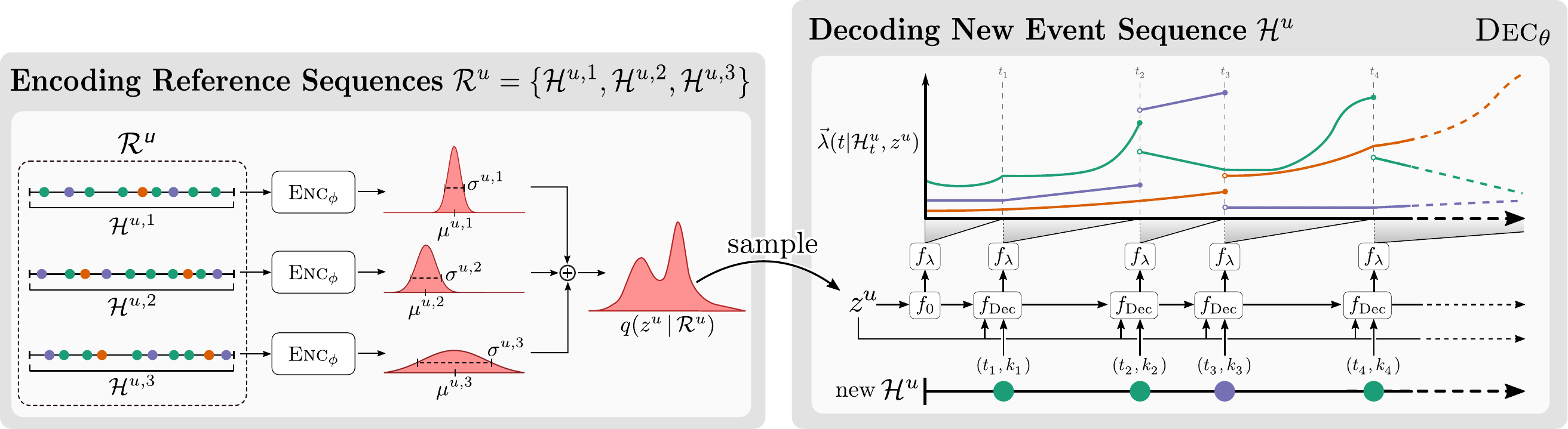}
    \caption{On the left is the encoding process for three reference sequences from user $u$ that belong to the reference set $\mathcal{R}^u$. This results in the approximate posterior mixture distribution $q(z^u\,|\,\mathcal{R}^u)$ that is then sampled and used in the decoding process on the right for a target sequence $\mathcal{H}^u$. $f_\lambda$ is the model specific transformation of hidden state $h_i$ and time $t \in (t_i, t_{i+1}]$ to intensity rates across marks.}
    \label{fig:enc_dec_overview}
\end{figure}

This section specifies the details for representing and estimating user embedding $z^u$, the underlying data generating process for sequential event data, how existing neural MTPP models can be extended to become personalized by incorporating user embeddings, as well as the loss function being optimized. Figure~\ref{fig:enc_dec_overview} shows operational diagrams of the encoding and decoding processes.

\paragraph{Encoding User Embeddings}
The user embedding~$z^u$ in Eq.~\ref{eq:stoch_rnn} allows us to personalize predictions for a given user~$u$.
$z^u$ is a real-valued vector, and for our results the dimensionality ranges from 32 to 64 depending on the dataset (see supplement for more information). The vector $z^u$ can be interpreted as the sequence and user-specific dynamics for a single history of events.
We infer~$z^u$ from a reference set~$\mathcal R^u=\{\mathcal{H}^{u,1}, \dots, \mathcal{H}^{u,n^u}\}$ of $n^u$ sequences that we have already observed from the same user.
This leads to two complications:
first, the amount of data in each reference set~$\mathcal R^u$ is much smaller than the whole training set of sequences from all users;
second, learning an individual user embedding~$z^u$ for thousands of users would be expensive.
We address both complications by an approximate Bayesian treatment of~$z^u$ via amortized variational inference (amortized VI)~\citep{KW2014,rezende2014stochastic,zhang2019advances}.

The typically small amount of data in each reference set~$\mathcal R^u$ motivates a Bayesian treatment via a probabilistic generative process with a prior $p(z^u)$ and the likelihood in Eq.~\ref{eq:stoch_rnn}.
For simplicity, we assume a standard normal prior: $z^u\sim\mathcal N(0,I)$.
Bayesian inference seeks the posterior probability $p(z^u | \mathcal R^u)$.
As finding the true posterior is computationally infeasible, VI approximates the posterior with a parameterized variational distribution $q(z^u|\mathcal R^u)$ by minimizing the Kullback-Leibler divergence from the true posterior to~$q$. 
The inferred approximate posterior $q(z^u|\mathcal R^u)$ can then be used to sample a user embedding~$z^u$ for a personalized prediction, i.e.,
\begin{align}\label{eq:pz_qz}
    z^u &\sim \begin{cases}
        \mathcal N(0, I) & \text{for unconditional generation;} \\
        q(z^u \,|\, \mathcal R^u) & \text{for personalized prediction.}
    \end{cases}
\end{align}
In principle, one could fit an individual variational distribution $q(z^u|\mathcal R^u)$ for each user~$u$.
As this would be expensive, we instead use
amortized VI \citep{KW2014}
We first model $q(z^u|\mathcal R^u)$ by a mixture of experts \citep{shi2019variational} where each expert $q(z^u | \mathcal H^u)$ is conditioned only on a single reference sequence $\mathcal H^{u} \in \mathcal R^u$, 
\vspace{-0.3em}\begin{align}\label{eq:moe}
    q(z^u \,|\, \mathcal R^u) = \textstyle\frac{1}{n^u}\sum_{i=1}^{n^u} q(z^u \,|\, \mathcal H^{u,i}).
\end{align}
$q(z^u \,|\, \mathcal R^u)$ represents the various modes of dynamics for a given user $u$ as defined by their past sequences $\mathcal R^u$. Further, each expert distribution $q(z^u | \mathcal H^u)$ is a fully factorized normal distribution where the means~$\mu$ and variances~$\sigma^2$  are further parameterized by an encoder neural network $\textsc{Enc}_\phi$ with parameters~$\phi$ that are shared across all users,
\begin{align}\label{eq:z_expert}
    q(z^u \,|\, \mathcal H^u) = \mathcal N\big(z^u; \mu, \text{diag}(\sigma^2)\big)
    \qquad\text{where}\qquad
    (\mu, \log\sigma) = \textsc{Enc}_\phi(\mathcal{H}^u).
\end{align}
$\textsc{Enc}_\phi$ contains a bidirectional RNN (more specifically gated recurrent units) that takes embedded times and marks of the reference sequence as inputs. The mark embeddings are learned and the time embeddings are continuous versions of the fixed positional embeddings %
\citep{cho2014properties,vaswani2017attention}. The last hidden states from both directions are concatenated and then linearly transformed to result in $\mu$ and $\log \sigma$. Precise details for this process can be found in the supplement.%

Eqs.~\ref{eq:moe}-\ref{eq:z_expert} specify the variational family. 
We optimize over the variational parameters~$\phi$ using standard methods for Black Box VI \citep{blei2017variational,zhang2019advances}, i.e., by stochastic maximization of the evidence lower bound. %

\paragraph{Distributions for Events}
Marked temporal point processes (MTPP) are a broad class of processes used for modeling seqeunces of events $\mathcal{H}$. 
A common method to fully characterize a MTPP is through an intensity function, 
\begin{equation}
\lambda(t|\mathcal{H}_t) = \lim_{\delta \downarrow 0} \textstyle \frac{1}{\delta}P\big(|\mathcal{H}_{t+\delta}| - |\mathcal{H}_t| = 1\;\big|\; \mathcal{H}_t\big),
\label{eq:overall_intesnity}
\end{equation}
where $|\mathcal{H}_t|$ counts the number of events up to time $t$. 
The intensity function measures the instantaneous rate of occurrence for events at time $t$, conditioned on the history up until time $t$.
Mark-specific intensity functions 
are defined as the product of the overall intensity function and the conditional categorical distribution over marks, $\lambda_k(t|\mathcal{H}_t) = p(k|t,\mathcal{H}_t)\lambda(t|\mathcal{H}_t)$. 
We denote the vector of rates over all marks as $\vv{\lambda}(t|\mathcal{H}_t)$. 
The log-likelihood of a  sequence $\mathcal{H}$ works out to be
\begin{equation}
\log p(\mathcal{H})  = \textstyle \sum_{i=1}^{|\mathcal{H}|} \log \lambda_{k_i}(t_i|\mathcal{H}_{t_i}) - \int_{0}^{T} \lambda(t|\mathcal{H}_t)dt.
\label{eq:ll}
\end{equation}
Intuitively, the summation in the first term rewards the model when the intensity values are high for the actual observed marks within the history, whereas the negative integral term penalizes high overall intensity values when there is no event. 

Intensity functions have been parameterized in both simple forms for interpretability as well as with neural networks for flexibility. 
Our proposed approach can in principle be used to add personalization to most existing neural MTPP models. 
We selected two of the most well-known and widely-cited such models to serve as base architectures which we extend for personalization as described in Sec.~2:
\begin{itemize}[leftmargin=12pt,itemsep=1pt]
    \item The \emph{Recurrent Marked Temporal Point Process} (RMTPP)~\citep{du2016recurrent}, which  parameterizes the intensity function explicitly as a piece-wise exponentially decaying rate: 
    $\overrightarrow{\lambda}^{\text{RMTPP}}(t|\mathcal{H}_t)=\exp\{Wh_i + w(t-t_i) + b\}$, where $t_i < t \leq t_{i+1}$, and $W$, $w$, and $b$ are learnable parameters. For this model, $f_\text{Dec}$ from Eq.~\ref{eq:stoch_rnn} is a gated recurrent unit (GRU). 
    \item The \emph{Neural Hawkes Process} (NHP)~\citep{mei2017neural}, which describes a procedure to obtain  an interpolated hidden state $h(t)$, defines
    $\overrightarrow{\lambda}^{\text{NHP}}(t|\mathcal{H}_t)=\text{softplus}(Wh_i(t))$ where $W$ is a learnable matrix with $t_i < t \leq t_{i+1}$.
    and has $f_\text{Dec}$ from Eq.~\ref{eq:stoch_rnn} be a continuous-time %
    LSTM unit.
\end{itemize}
By incorporating $z^u$ as specified in Eq.~\ref{eq:stoch_rnn}, $\overrightarrow{\lambda}(t|\mathcal{H}^u_t)$ becomes $\overrightarrow{\lambda}(t|\mathcal{H}^u_t,z^u)$. Note that by defining $\overrightarrow{\lambda}(t|\mathcal{H}^u_t,z^u)$ and $f_\text{Dec}$, we effectively define $p(t_{i+1},k_{i+1}\,|\,h_i,z^u)$ in Eq.~\ref{eq:stoch_rnn}, as well as the general decoding process, $\textsc{Dec}_\theta$.

All MTPP models can be sampled via a thinning procedure \citep{ogata1981lewis}, if not directly. Similarly, the integral in Eq.~\ref{eq:ll} can be computed by Monte-Carlo estimation, if not analytically. In our experiments, we perform the former for all models for consistency. More precise details on this can be found in the supplement.

\paragraph{Optimization} 

The objective function for the proposed personalized neural MTPP models is the $\beta-\text{VAE}$ objective \citep{higgins2017beta} which is defined for a single target sequence $\mathcal{H}^u$ as:
\begin{align}
\label{eq:ELBO}
    {\cal L}_\beta(\phi, \theta; \mathcal{H}^u)  =  {\mathbb{E}}_{q_\phi(z^u|{\cal R}^u)} [\log p_\theta({\cal H}^u|z^u)] -\beta {\rm KL}(q_\phi(z^u|{\cal R}^u)|| p(z^u)),
\end{align}
which the right-hand side is a variant of what is known as the evidence lower bound (ELBO).
The expectation is estimated with a Monte-Carlo estimate. During training, a single sample for the estimate turned out to be sufficient, whereas during testing we utilized 5 samples to reduce variance.

\section{Experimental Results}

To measure the effectiveness of this framework in general, using the NHP and RMTPP base models we trained each in their standard configuration (i.e., a decoder-only setup) 
and in the proposed variational mixture-of-experts setup (referred to as MoE-NHP and MoE-RMTPP).\footnote{Our source code for modeling and experiments can be found at the following repository: \url{https://github.com/ajboyd2/vae_mpp}.} We rated these models  on their held-out log-likelihood values, next event predictions, and user/source identification capabilities as described below.\footnote{Sampling quality was also evaluated. Details and results can be found in the Supplement.} Furthermore, all tests were conducted using sequences from new users to emphasize the added ability to adapt to new sources.

Models were trained by minimizing Eq.~\ref{eq:ll} and Eq.~\ref{eq:ELBO}, averaged over training sequences, for the decoder-only and MoE variants respectively via the Adam optimizer with default hyperparameters \citep{kingma2014adam} and a learning rate of 0.001. 
A linear warm-up schedule for the learning rate over the first training epoch was used as it led to more stable training across runs. We also performed cyclical annealing on $\beta$ in Eq.~\ref{eq:ELBO} from 0 to $0.001$ with a period of 20\% of an epoch to help prevent the posterior distribution from collapsing to the prior \citep{liu2019cyclical}.

\begin{table}[t!]
    \caption{Statistics for the four datasets. Columns (left to right) are: total time window $T$ for every sequence;   number $K$ of unique marks;   mean sequence length $|\mathcal{H}|$;   mean number $|\mathcal{R}^u|$ of sequences per user;   total number of sequences and number of unique users in training/validation/test splits. 
    }
    \label{tab:datasets}
    \centering
    \begin{tabular}{llrrrrrrrrrr}
    \hline
    & & & Mean & Mean & \multicolumn{3}{c}{\# Sequences} & \hspace{-1em} &  \multicolumn{3}{c}{\# Unique Users} \\
    Dataset & \mc{$T$} & \mc{$K$} & \mc{$|\mathcal{H}|$} & \mc{$|\mathcal{R}^u|$} & Train & Valid & Test & \hspace{-1em} &  Train & Valid & Test \\
    \hline\hline
    Meme   & 1 Week  & 5000 & 23.4 & 6.9 & 271K & 9K  & 21K &\hspace{-1em} & 31K & 1K & 3K \\
    Reddit & 1 Week  & 1000 & 65.2 & 4.9 & 343K & 15K & 34K &\hspace{-1em} & 49K & 3K & 7K \\
    Amazon & 1 Month & 737  & 10.7 & 4.0 & 262K & 8K  & 20K &\hspace{-1em} & 28K & 2K & 5K \\
    LastFM & 1 Day   & 15   & 45.6 & 347 & 289K & 15K & 49K &\hspace{-1em} & 833 & 44 & 105\\
    \hline
    \end{tabular}
\end{table}

\subsection{Datasets}
All models were trained and evaluated on four real-world datasets (see Table~\ref{tab:datasets}).
The {\bf MemeTracker} dataset \citep{snapnets} relates to common phrases (memes).
We defined the meme as the ``user'' and the website it was posted to as the mark. The mark vocabulary is the set of top 5000 websites by volume of utterances. Sequences were defined as one-week-long chunks spanning August 2008 to April 2009, and event times were measured in hours.   The {\bf Reddit comments} dataset \citep{baumgartner2020pushshift} relates to user-comments  on posts  in the social media site \url{reddit.com}. One month of data (October 2018) was used to extract user  sequences, and the mark vocabulary was defined as the top 1000 communities (subredits) by comment volume.  The month was divided into multiple sequences consisting of week-long windows per user, with event times  in units of hours. {\bf Amazon Reviews} \citep{ni2019justifying} consists of timestamped product reviews  between May 1996 and October 2018, with marks  defined as 737  different product categories. User sequences were defined as 4-week windows with event times in units of days (with a small amount of uniformly distributed noise to avoid having multiple events co-occur). The 4th dataset, {\bf LastFM} \citep{Celma:Springer2010}, has time-stamped records of  songs listened to (both artists and track names) for nearly 1,000 users on the \url{last.fm} website. Marks were defined as one of 15 possible genres for a song, via  the \url{discogs.com} API. User sequences corresponded to 1 day of listening in units of hours. For all datasets event times were calculated relative to the start-time of a user sequence.  
All datasets were filtered to include sequences with at least five events and no more than 200. Training, validation, and test sets were split so that there were no users in common between them.

\subsection{Results}

\paragraph{Training Data Size Ablation}
We first investigate differences in predictive performance between the two proposed Mixture of Experts~(MoE) models and their decoder-only counterparts, as a function of the size of the training data.
We therefore trained each model on various subsets of the training data, using $10\%$, $20\%$, $30\%$, $50\%$, $70\%$, $90\%$, and $100\%$ of the full training set.
For efficiency reasons, we generated these trained models using curriculum learning, i.e., we first trained each model on the $10\%$ subset until convergence, then added $10\%$ more training points and trained on the resulting $20\%$ subset, and so on.
Convergence on each subset was determined when validation log-likelihood improved by less than $0.1$.
The training subsets were generated via random sampling of users.  All models were evaluated on the same fixed-size test dataset.

\begin{figure}[t!]
  \includegraphics[width=\linewidth]{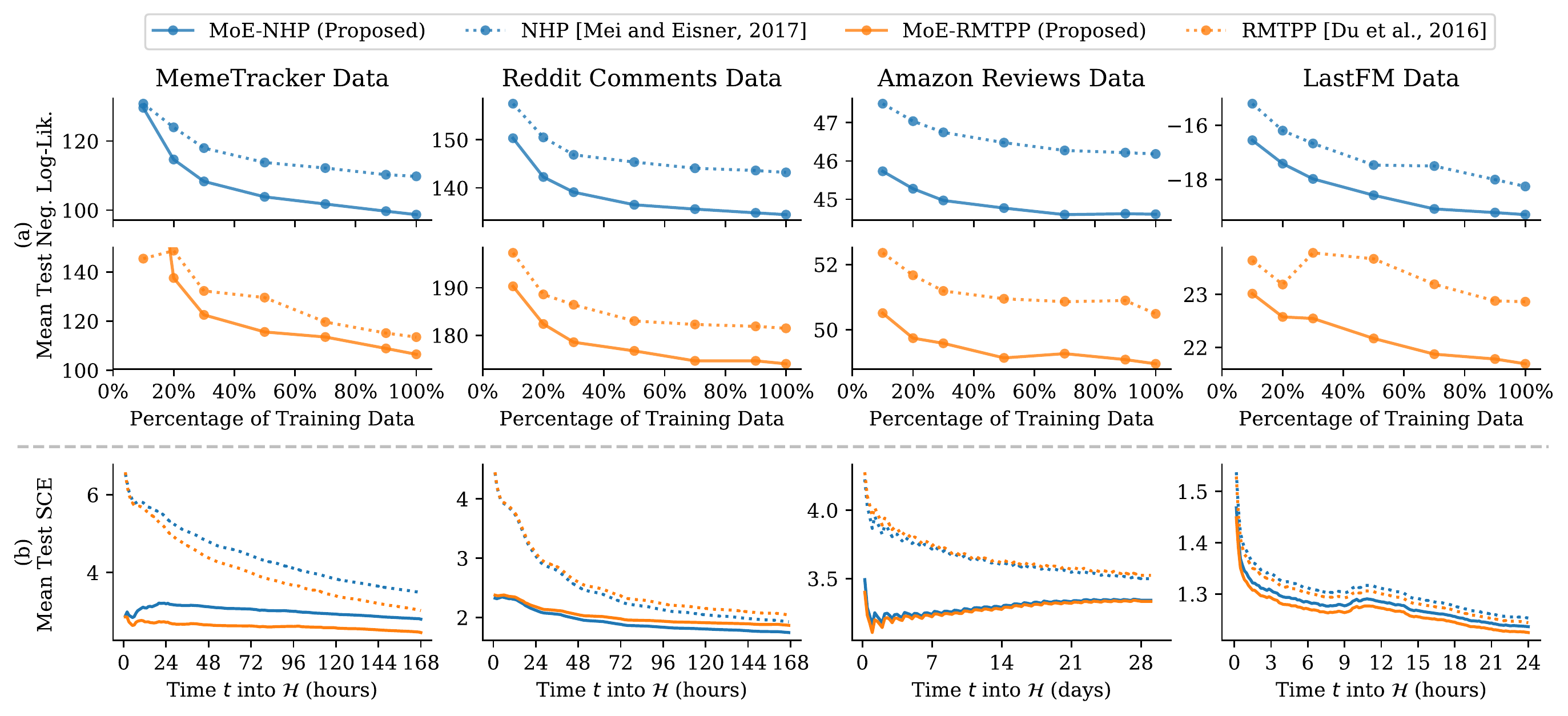}
  \caption{(a) Mean test negative log-likelihood performance for models trained at varying percentages of the training data across the four datasets and (b) mean test cross entropy up to time $t$ (see Eq.~\ref{eq:ll_broken}) for models trained on 100\% of the training data (lower is better for both). Results for NHP-based models are shown in blue, RMTPP models are orange, decoder-only models are dotted lines, and the proposed mixture-of-experts models are solid lines.}
  \label{fig:performance}
\end{figure}

The results can be seen in Figure~\ref{fig:performance}(a). The proposed MoE models (solid lines) systematically yield better  predictions in terms of  test log-likelihood over their non-MoE counterparts (dotted lines).
This trend suggests that our personalization scheme could benefit most, if not all, neural MTPP models and that these benefits appear for even small amounts of training data.

\paragraph{Likelihood Over Time}
Having seen that the proposed MoE models have better predictive log-likelihoods than their decoder-only counterparts, we now investigate where exactly they achieve these performance gains.
Using $\lambda_k(t|\mathcal{H}_t) = p(k|t,\mathcal{H}_t)\lambda(t|\mathcal{H}_t)$, we can factor Eq.~\ref{eq:ll} as follows:
\begin{align}\label{eq:ll_broken}
    -\log p(\mathcal{H})   = |\mathcal{H}|\left({\text{SCE}}(\mathcal{H}) + {\text{PP}}^{+}(\mathcal{H})\right) + T\;\!{\text{PP}}^{-}(\mathcal{H}) 
\end{align}
where ${\text{SCE}}(\mathcal{H}_t)   \equiv \frac{-1}{|\mathcal{H}_t|} \sum_{i=1}^{|\mathcal{H}_t|} \log p(k_i| t_i,\mathcal{H}_{t_i}),$ is the average cross entropy of a sequential, non-continuous time based, classification model,
and 
${\text{PP}}^{+}(\mathcal{H}_t)   \equiv \frac{-1}{|\mathcal{H}_t|} \sum_{i=1}^{|\mathcal{H}_t|} \log \lambda(t_i |\mathcal{H}_{t_i}),  
{\text{PP}}^{-}(\mathcal{H}_t)   \equiv \frac{1}{t}\int_{0}^{t} \lambda(\tau|\mathcal{H}_{\tau}) d\tau $ respectively represent the average positive and negative evidence of a sequence, ignoring the associated marks (the two together make up the terms in the log-likelihood of a non-marked temporal point process). 
All of these terms can assist in identifying issues within MTPPs, especially when investigated as a function of $t$. We presume that most of the heterogeneity between sequences in datasets resides in the categorical distributions of marks. Here we focus our analysis on the ${\text{SCE}}$ term---see the supplementary material for discussion of other terms.

Figure~\ref{fig:performance}(b) shows average ${\text{SCE}}$ values over time. The decoder-only models (dotted) tend to have high ${\text{SCE}}$ values  near the beginning of sequences (left side of x-axis), as  the model adapts to the type of sequence and user it is making predictions for. 
In contrast, the MoE models (solid) have  much lower ${\text{SCE}}$ near the beginning of sequences, i.e., are making significantly better categorical predictions for marks %
early on. The decoder-only models gradually approach the performance of the MoE models over time, but never close the gap, indicating the user information (via $z_u$) provides significant benefit in mark prediction that is difficult for an RNN decoder model to learn from the sequence itself.

\paragraph{Ranking of Next Event Predictions}
One use case of MTPP models is to predict what a user will do next and when they will do it during an ongoing sequence.
As an example scenario, we conditioned the models on a prefix $\mathcal{H}^u_{t_{10}}$ of the first $10$ events in each test sequence and evaluated the predictive performance for the next event $(t_{11},k_{11})$.
Predicted times and marks were estimated by marginalizing over marks and times respectively
to minimize Bayes risk, similar to \cite{mei2017neural}---see the supplement for details.
The choice of $10$ events in the prefix was made to simulate making predictions early in a sequence, where there is still a good deal of variability for the next event. 

The predicted marks were evaluated by the ranking of the true mark's predicted probability, averaged over all sequences in the test set for each dataset. 
Table~\ref{tab:experiment_results}(a) shows our results.
MoE models achieve superior performance than the other models on all datasets, with particularly large gains for the Meme and Reddit data. %
The results for predicted times were not found to be consistently different between the MoE and non-personalized models. One possible reason for this might be that there is not a strong user-specific signal in event timing information that cannot already be detected in the sequence being decoded, thus causing the personalization models to focus more on event types than on times. In terms of predictive performance, it appears that the biggest benefit of personalization is more accurate predictions of marks rather than times. All results for predicted times can be found in the supplement.

\paragraph{Source Identification}
The personalized MoE models naturally lend themselves to being able to detect anomalous events for a given user.
We evaluate how well the models can identify the source of a given event sequence.
As discussed in Sec.~2, the MoE models were designed such that conditioning on a user embedding, subsequent sequences from the same user would be more likely (hence being personalized), and in turn sequences from different users would be less likely. %
We assessed this behavior on all four datasets by randomly selecting target sequences and pairing them up with one reference sequence from the same user and one reference sequence from a different user. After conditioning on the reference sequences, the models evaluated the likelihood of the target sequence and the two results were ranked amongst each other. This was done 5,000 times for each test dataset (10,000 total likelihood calculations each). The target sequences were truncated to 10 events in order to simulate identifying users from only a portion of events into a session.

The MoE-NHP and MoE-RMTPP models were compared against a Bayesian Poisson process baseline with a Gamma prior where the prior expected value was determined by the MLE of the training data and the strength of the prior was tuned on the validation set. Each target sequence was then evaluated on the posterior distribution conditioned on the reference sequence---more details can be found in the supplement. The results in  Table~\ref{tab:experiment_results}(b) show that both of  the MoE models have significantly lower error rates than the baseline on this task, across all four datasets.

\begin{table}[]
\caption{(a) Mean predicted mark rank for the proposed MoE-NHP and MoE-RMTPP models versus the non-personalized NHP and RMTPP models. More precision is shown for datasets with fewer total marks. (b) Source identification error rates for MoE-NHP, MoE-RMTPP, and Bayesian Poisson process baseline across the four datasets. For both, lower percentages are better and the best performance within a grouping is bolded. In all categories, one of our proposed models is the best.}
\label{tab:experiment_results}
\centering
\resizebox{\columnwidth}{!}{
\begin{tabular}{@{$\;$}l@{$\;$}c@{$\;\;$}c@{$\;\;$}c@{$\;\;$}c@{}c@{$\;\;$}l@{$\;$}c@{$\;\;$}c@{$\;\;$}c@{}}
\cline{1-5}\cline{7-10}
  \multicolumn{5}{c}{(a) Mean Predicted Rank ($\hat k_{11}$)} & \hspace{1em} & \multicolumn{4}{c}{(b) Source Identification Error Rate} \\
Dataset & MoE-NHP & NHP & MoE-RMTPP & RMTPP & \hspace{-1em} & $\;$Dataset & MoE-NHP & MoE-RMTPP & Bayes PP\\
\cline{1-5}\cline{7-10} \multicolumn{9}{c}{} \\[-0.8\normalbaselineskip] \cline{1-5}\cline{7-10}
Meme & \textbf{225} & 283 & \textbf{143} & 159  & \hspace{-1em} & $\;$Meme & 5.04\% & \textbf{4.64\%} & 9.18\% \\
Reddit & \textbf{19.4} & 35.2 & \textbf{20.0} & 35.6  & \hspace{-1em} & $\;$Reddit & \textbf{1.38\%} & 2.02\% & 3.24\% \\
Amazon & \textbf{21.6} & 22.8 & \textbf{21.6} & 22.9  & \hspace{-1em} & $\;$Amazon & \textbf{8.34\%} & 9.94\% & 12.88\% \\
LastFM & \textbf{2.02} & 2.04 & \textbf{2.01} & 2.04  & \hspace{-1em} & $\;$LastFM & \textbf{28.44\%} & 33.90\% & 39.94\% \\
\cline{1-5}\cline{7-10}
\end{tabular}

}
\end{table}

\section{Related Work}

There has been a significant amount of recent work on neural MTPPs that has focused on incorporating additional information outside of the event sequence itself, such as dense time-series information \citep{xiao2017modeling}.
There have also been advances in bypassing computational issues involved in evaluating integrals in the MTPP likelihood, e.g., via direct modeling of cumulative hazard functions \citep{omi2019fully} or via the use of normalizing flows \citep{shchur2019intensity}. 
Other approaches have avoided these issues through alternative objectives à la 
reinforcement learning \citep{li2018learning,upadhyay2018deep,zhu2019reinforcement} or adversarial training \citep{xiao2018learning,xiao2017wasserstein}.

In the context of user-specific sequence models,
\citet{vassoy2019time} introduced MTPPs with learned user embeddings for applications with many sequences per user to personalize next step recommendations. In contrast, our work focuses on forecasting entire future trajectories of user activity in situations with few sequences per user. Additionally, we concern ourselves with predicting for users that were not necessarily present during model fitting.

Sequential VAE models \citep{bayer2014learning,chung2015recurrent,gan2015deep} have been developed for discrete-time and dense data such as text~\citep{bowman2016generating,yang2017improved,schmidt2019autoregressive} or video \citep{yingzhen2018disentangled,denton2018stochastic} with applications in compression \citep{lombardo2019deep} and control \citep{chua2018deep,ha2018recurrent}. However, none of these works have been extended to the continuous-time and sparse data domain that we encounter in event sequence modeling.

\section{Conclusion}

We address the problem of personalization in discrete-event continuous time sequences by proposing a new framework that integrates with existing neural MTPP models based on a variational mixture-of-experts autoencoder approach. 
Our analysis shows that, across four different real world datasets, the introduction of personalization via our proposed framework improved predictive performance on metrics such as held-out log-likelihood, next event prediction, and source identification.
Our approach opens up possibilities for various personalized tasks that were not readily feasible before with prior neural MTPP works. Such tasks include personalized sequential recommendations, demand forecasting, and source-wise anomaly detection.

It should be noted, however, that the fact that the $\beta$-VAE approach, with $\beta < 1$, was necessary showed that our approach suffers from similar problems as sequential VAEs for text generation \cite{bowman2016generating}. This makes the learned representations less useful in practice and opens up the question on which models are better suited for representation learning of discrete event sequences. We foresee future work tackling this issue through potentially several avenues: better model architectures (for either the encoder or decoder), better multi-modal latent modeling mechanisms instead of the mixture-of-experts approach, or even more appropriate priors that better match the underlying sequence dynamics.

\section*{Acknowledgements}

This material is based upon work supported in part by the National Science Foundation under grant numbers 1633631, 1928718, 2003237, and 2007719; by the National Science Foundation Graduate Research Fellowship under grant number DGE-1839285; by the Center for Statistics and Applications in Forensic Evidence (CSAFE) through Cooperative Agreement 70NANB20H019 between NIST and Iowa State University, which includes activities carried out at University of California Irvine; by the Defense Advanced Research Projects Agency (DARPA) under Contract No. HR001120C0021; by an Intel grant; and by a Qualcomm Faculty Award. Any opinion, findings, and conclusions or recommendations expressed in this material are those of the authors and do not necessarily reflect the views of the National Science Foundation, nor do they reflect the views of DARPA.

Additional revenues related to this work include: employment with the Walt Disney Company and Google; research funding from NIH, NASA, NIST, PCORI, SAP and eBay; honoraria from General Motors; consulting income from Toshiba; and internships at Workday Incorporated, NVIDIA, and Microsoft Research.

\section*{Broader Impacts}

While many of the successful and highly-visible applications of machine learning are in classification and regression, there are a broad range of applications that don't naturally fit into these categories and that can potentially benefit significantly from machine learning approaches. In particular, in this paper we focus on continuous-time event data, which is  very common in real-world applications but has not yet seen significant attention from the ML research community. There are multiple important problems in society where such data is common and that could benefit from the development of better predictive and simulation, including:
\begin{itemize}[leftmargin=12pt]
    \item {\bf Education:} Understanding of individual learning habits of students, especially in online educational programs, could improve and allow for more personalized curricula.
    \item {\bf Medicine:} Customized tracking and predictions of medical events could save lives and improve patients' quality of living. %
    \item {\bf Behavioral Models:} Person-specific simulations of their behavior can lead to better systematic understandings of people's social activities and actions in day-to-day lives.
    \item {\bf Cybersecurity:} Through the user identification capabilities, our work could aid in cyber-security applications for the purposes of identifying fraud detection and identify theft. %
\end{itemize} 

Another potential positive broad impact of the work, is that by utilizing amortized VI, our methods do not require further costly training or fine-tuning to accommodate new users, which can potentially produce energy savings and lessen environmental impact in a production setting.
 
 On the other hand, as with many machine learning technologies, there is also always the potential for negative impact from a societal perspective. For example, more accurate individualized models for user-generated data could be used in a negative fashion  for applications such as surveillance (e.g., to monitor and negatively impact individuals in protected groups).
In addition, better predictions and recommendations for products and services, through explicitly conditioning on prior behavior from a user,  could potentially further worsen existing privacy concerns. %

\bibliography{point_process}
\newpage
\appendix

\section{Model Details}

\subsection{Encoder Details}

Below are precise details concerning the steps to encode a single sequence, $\mathcal{H}^u$, to describe its associated component distribution, $q(z^u\,|\,\mathcal{H}^u)$. 

\paragraph{Temporal Embedding}

The encoder contains a bidirectional RNN that accepts as input for each step the embedded mark vector, $\mathbf{k}_i$, and the embedded time vector, $\mathbf{t}_i$, for a given event in the sequence being encoded. The embedding for the mark is a standard, learned embedding. The embedding for the time is a fixed transformation that converts a single time $t$ into a $d_{\text{time}}$-dimensional vector. The specific temporal embedding function used in the models trained is defined as follows:

\begin{equation}
\begin{split}
\Phi(t) = [\sin(\alpha(t-t_i)); \cos(\alpha(t-t_i))] \\
\text{ for } t_i < t \leq t_{i+1}, \\
\end{split}
\end{equation}
where $t_i$ is the latest event time that the model has conditioned on that is less than the time $t$ being embedded, $\alpha$ is a fixed $\frac{d_{\text{time}}}{2}$-dimensional vector with the $j$th element being $\alpha_j = \exp\{-j\log(T_{max}) / d_\text{time}\}$ with $T_{max}$ being the maximum difference of consecutive times for a given dataset. This transformation is the same as the positional embeddings from \cite{vaswani2017attention} adapted to continuous times and can be seen as a simplified version of \cite{xu2019self}. This form was chosen to have a dense representation of time that safely generalizes to new time values. 

\paragraph{Encoding Events}

As mentioned previously, the embedded times and marks are used as inputs to a bidirectional RNN. More precisely:
\begin{align}
\overrightarrow{h}_i & = f_{\overrightarrow{\text{Enc}}}(\overrightarrow{h}_{i-1}, [\mathbf{t}_i; \mathbf{k}_i]), \\
\overleftarrow{h}_i & = f_{\overleftarrow{\text{Enc}}}(\overleftarrow{h}_{i+1}, [\mathbf{t}_i; \mathbf{k}_i]),
\end{align}
for $i=1,\dots,|\mathcal{H}^u|$ where $\overrightarrow{h}_0$ and $\overleftarrow{h}_{|\mathcal{H}^u|+1}$ are learned, and $f_{\overrightarrow{\text{Enc}}}$ and $f_{\overleftarrow{\text{Enc}}}$ are recurrent units (in our case, GRUs).

The information in the reference sequence is summarized by concatenating the last hidden states from each direction. We will denote that as $h=[\overrightarrow{h}_{|\mathcal{H}^u|};\overleftarrow{h}_1]$. This vector is then used to compute the sufficient statistics for the component distribution via:
\begin{align}
\mu = f_\mu(h) \;\;\text{ and } \;\; \log \sigma = f_\sigma(h)
\end{align}
where in our implementation, $f_\mu(h) = W_\mu h + b_\mu$ and $f_\sigma(h)=W_\sigma h + b_\sigma$ for learnable matrices $W_\mu$ and $W_\sigma$ and learnable bias vectors $b_\mu$ and $b_\sigma$.

\subsection{Model Visualizations}

\begin{figure}[t]
\begin{center}
\includegraphics[width=0.8\columnwidth]{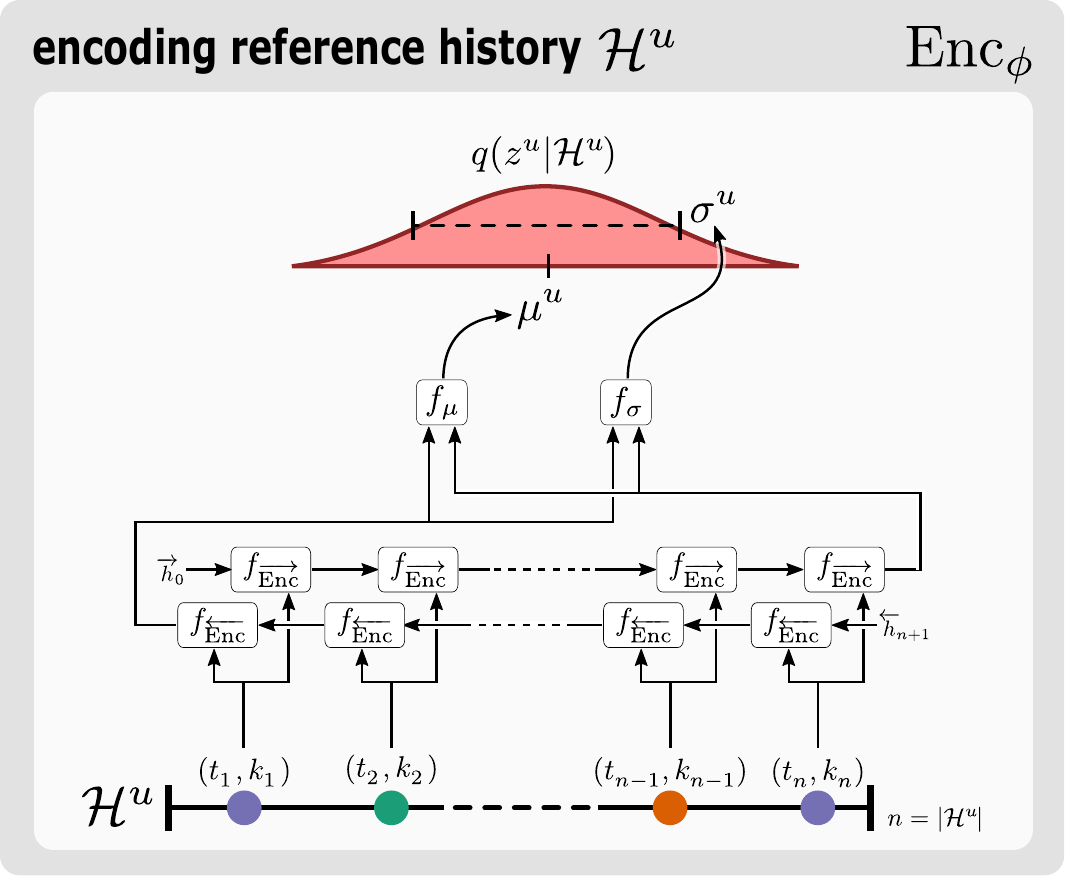}
\end{center}
\caption{A detailed operational diagram of the encoding process. A single sequence $\mathcal{H}^{u}$ is being encoded to compute the sufficient statistics, $\mu^{u}$ and $\sigma^{u}$, for a single component/mixture, $q(z^u\,|\,\mathcal{H}^{u})$.}
\label{fig:enc}
\end{figure}

\begin{figure}[t]
\begin{center}
\includegraphics[width=\columnwidth]{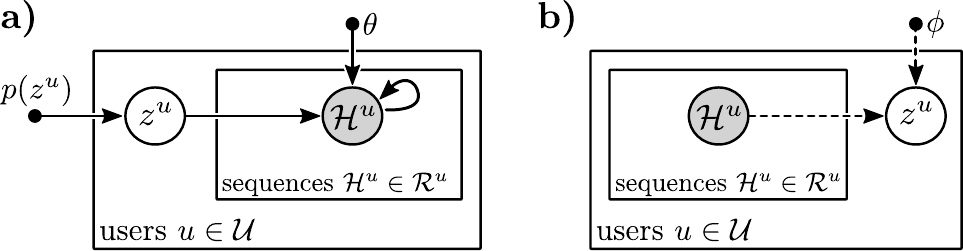}
\end{center}
\caption{Graphical model of the proposed VAE for personalized point processes with 
a) being the generative model and 
b) being the inference model. $\mathcal{U}$ is the set of all possible users for a given dataset.}
\label{fig:graphical-model}
\end{figure}

An operational diagram of the encoder, as described previously, can be found in Figure~\ref{fig:enc}. A graphical model representation of the personalized neural MTPP framework can be found in Figure~\ref{fig:graphical-model}.

\subsection{Sampling Sequences}

One convenient property of point processes is that given multiple point processes, the superposition of them is also a valid point process. Furthermore, the intensity function of the superposition point process is simply equal to the sum of intensity functions that make it up (e.g., for the combination of two point processes: $\lambda(t|\mathcal{H}_t) = \lambda_1(t|\mathcal{H}_t) + \lambda_2(t|\mathcal{H}_t)$). This is why $\sum_{k=1}^K \lambda_k(t|\mathcal{H}_t) = \lambda(t|\mathcal{H}_t)$ for MTPPs, as events for each mark can be seen as coming from their own point process. Note that no assumption of independence was made for this property to hold true.

We utilize this property in a thinning procedure to sample from arbitrary point processes. If we let $\lambda^* > \lambda(t|\mathcal{H}_t)$ for all $t \in [0,T)$, then to sample from $\lambda(t|\mathcal{H}_t)$ requires sampling a time, $t^*$, from a Poisson point process with constant rate $\lambda^*$, then randomly accepting that point as originating from the model with probability $\frac{\lambda(t^*|\mathcal{H}_t)}{\lambda^*}$. If it is accepted, then the mark is determined by sampling from a categorical distribution with probabilities equal to $\frac{\lambda_k(t^*|\mathcal{H}_t)}{\lambda(t^*|\mathcal{H}_t)}$ for $k \in \{1,\dots, K\}$. The time and mark are then appended to $\mathcal{H}$ and the procedure continues until candidate times are sampled outside of a pre-specified time window. Note that, should we want to condition on a portion of a history, $\mathcal{H}_c$, and sample future trajectories, then the intensity function is conditioned on $\mathcal{H}_c$ and the only candidate times considered are $t \in [c,T)$. 

For this thinning procedure to be valid, $\lambda^*$ must dominate all intensity values estimated by the model; however, this can be difficult to ensure prior to generating a sample due to the intensity function changing in response to new events. As such, to ensure that a sample was validly generated one must check the intensity function conditioned on the sample at various times in the time window and validate that it is less than $\lambda^*$. All samples that we generated had this check done at 1,000 different times that were uniformly sampled across the time window. Should this condition not hold for at least one point in the interval, then $\lambda^*$ is increased and a new sample is generated and subsequently validated.

\section{Experimental Details}

This section pertains to more precise and specific details concerning our experimental findings, as well as some additional results that were omitted from the main text due to space.

\subsection{Experiment Hyperparameters}

Below are descriptions that list all of the hyperparameters set throughout our training and experiments, such as the specific sizes for model parameters or the number of samples used when approximating integrals.

\paragraph{Model Architecture}

Table~\ref{tab:model_hyper} contains model hyperparameters used for all of the experiments. The same hyperparameters were sufficient for models trained on all datasets; however, due to the difference in total unique marks (i.e. $K_\text{MemeTracker}=5000$, $K_\text{Reddit}=1000, K_\text{Amazon}=737$, and $K_\text{LastFM}=15$) it turned out necessary to have the size of the amortized latent user embedding, $z_u$, be 64 for the MemeTracker data instead of 32 as used for the Reddit and Amazon data. Similarly, we found it necessary that the decoder hidden size was much larger for the models trained on the LastFM data due to different trends and patterns being harder to differentiate from the subset of marks present alone. These values were found from ablation studies using splits of the training data for each dataset. We suspect that since each dataset had a similar amount of observations, the models tended to require similar minimum capacities. 

For the MoE models, to save on parameters, the same mark embeddings were shared amongst the encoder and decoder.

\begin{table}[t!]
    \centering
    \caption{Descriptions and values of hyperparameters used for models trained on all of the datasets. The same decoder hyperparameters were used on all of the models, whereas the encoder hyperparameters were only used for the MoE variants.}
    \begin{tabular}{clcccc}
    \hline
    & & \multicolumn{4}{c}{Value Used} \\
    Section & Description & Meme & Reddit & Amazon & LastFM \\
    \hline\hline
    \multirow{4}{*}{Encoder} & Temporal Embedding Size & 64 & 64 & 64 & 64 \\
    & Mark Embedding Size & 32 & 32 & 32 & 32 \\
    & Encoder Hidden Size & 64 & 64 & 64 & 64 \\
    & Latent State Size & 64 & 32 & 32 & 32 \\
    \vspace{-0.5em} \\
    \multirow{2}{*}{Decoder} & Mark Embedding Size & 32 & 32 & 32 & 32 \\
    & Decoder Hidden Size & 64 & 64 & 64 & 256 \\
    \hline
    \end{tabular}
    \label{tab:model_hyper}
\end{table}

\paragraph{Approximations}

For training and experiments, there are a number of integrals that need to be computed which are not feasible in closed form. Thus, we must approximate them. All integrals and expectations are approximated via Monte-Carlo (MC) estimates with varying amounts of samples used.

In the log-likelihood of a sequence, Eq.~7, the term $\int_0^T \lambda(t|\mathcal{H}_t)dt$ uses 150 MC samples during training and 500 MC samples for evaluating held-out log-likelihood values for experiments. The exact approximation procedure for the log-likelihood can be found in \citet{mei2017neural}. Similarly, in the objective function for the MoE models, Eq.~8, the expectation term ${\mathbb{E}}_{q_\phi(z^u|{\cal R}^u)} [\log p_\theta({\cal H}^u|z^u)]$ uses a single MC sample drawn from $q_\phi(z^u|{\cal R}^u)$ for training, and 5 MC samples for evaluating held-out log-likelihood values for experiments.

When evaluating the integrals used for next event predictions, we used 10,000 samples where the sample points were shared across integrals for a single set of predictions in order to save on computation. This approach is the same as executed in \citet{mei2017neural}.

\subsection{Likelihood Over Time Analysis}

\begin{figure*}[t!]
  \includegraphics[width=\textwidth]{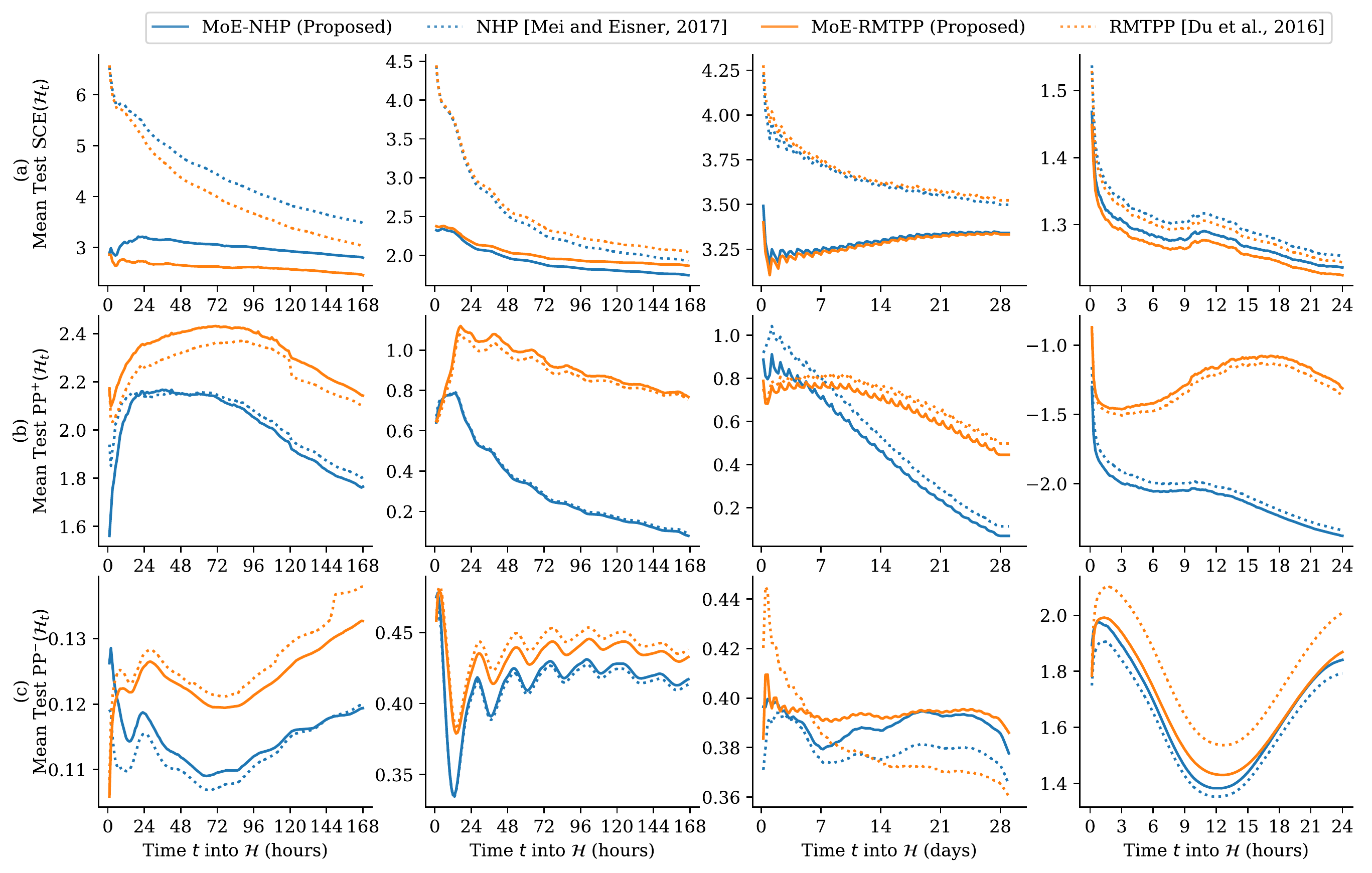}
  \caption{Mean values of ${\text{SCE}}, {\text{PP}}^+,$ and ${\text{PP}}^-$ (seen in rows (a), (b), and (c), respectively) for decoder-only (dotted lines) and MoE (solid lines) variants of NHP (blue lines) and RMTPP (orange lines) models with each column corresponding to results for the MemeTracker dataset, Reddit comments, Amazon reviews, and the LastFM dataset. In every plot, a lower value is better. Note that for the Amazon results, the periodic trends shown can most likely be attributed to the small uniform noise applied to the data to avoid multiple events co-occurring.}
  \label{fig:all_loss}
\end{figure*}

In the main text, we broke down the negative log-likelihood of a sequence $\mathcal{H}$ up to time $t$ into the normalized cross entropy ${\text{SCE}}(\mathcal{H}_t) = \frac{-1}{|\mathcal{H}_t|} \sum_{i=1}^{|\mathcal{H}_t|} \log p(k_i| t_i,\mathcal{H}_{t_i})$, the normalized positive point process evidence ${\text{PP}}^{+}(\mathcal{H}_t)   = \frac{-1}{|\mathcal{H}_t|} \sum_{i=1}^{|\mathcal{H}_t|} \log \lambda(t_i |\mathcal{H}_{t_i})$, and the normalized negative point process evidence ${\text{PP}}^{-}(\mathcal{H}_t) = \frac{1}{t}\int_{0}^{t} \lambda(\tau|\mathcal{H}_{\tau}) d\tau$. The first of which assesses the model's sequential classification performance and the latter two assesses the model's performances as a non-marked temporal point process (or in other words, how well it captures the time dynamics of the event data). More specifically, the positive evidence measures how well the model reports high intensity rates for all events when an event actually occurs, and the negative evidence quantifies how well the model estimates low intensity values during periods of no events occurring. These defined terms have all been normalized so that they may be compared across different amounts of time into a sequence. They have also been appropriately negated so that for each term, a lower value is desirable.

Results for all three terms across all four model variants and all four datasets can be seen in Figure~\ref{fig:all_loss}. We observe that the cross entropy for every dataset is superior for the proposed, personalized models compared to their decoder-only counterparts, especially near the beginning of the sequence where there is the most uncertainty. The story is not as consistent for the other two terms as it appears there is a trade-off between them. In most instances, an average set of lower $\text{PP}^{+}$ values results in higher $\text{PP}^{-}$ values when comparing MoE models to their decoder-only counterparts. 

From this, we would conclude that our proposed personalization scheme appears to consistently improve held-out log-likelihoods primarily due to better modeling of the sequential mark distributions, whereas it is a mixed bag for improving the distributions over event timings.

\subsection{Next Event Predictions} 

As described in \citet{mei2017neural}, we minimized the Bayes risk to determine decisions for what a predicted next time $\hat t_{i+1}$ and mark $\hat k_{i+1}$ would be after conditioning on a portion of a sequence $\mathcal{H}_{t_{i}} = [(t_1, k_1), \dots, (t_i, k_i)]$. 

For the former prediction ($\hat t_{i+1}$), we first note that the next event time $t_{i+1}$ has a density $p_{i+1}(t) = P(t_{i+1}=t\,|\,\mathcal{H}_{t_i})=\lambda(t|\mathcal{H}_t)\exp\left\{-\int_{t_i}^t \lambda(s|\mathcal{H}_s)ds\right\}$. We define the precitec time $\hat t_{i+1}$ as the expected time under $p_{i+1}(t)$, i.e., $\hat t_{i+1} = \mathbb{E}[t_{i+1}\,|\,\mathcal{H}_{t_{i+1}}]=\int^{\infty}_{t_i} tp_i(t) dt$.
With this definition, $\hat t_{i+1}$ minimizes the expected $L_2$ distance $\mathbb E_{p_{i+1}(t)}[(t - \hat t_{i+1})^2]$.

The latter prediction ($\hat k_{i+1}$) is computed similarly via $\hat k_{i+1} = \text{argmax}_k \int_{t_i}^\infty \frac{\lambda_k(t|\mathcal{H}_t)}{\lambda(t|\mathcal{H}_t)}p_i(t)dt$. Note that this prediction of $k_{i+1}$ does not condition on the predicted time. These can be viewed as marginal predictions.  

As mentioned previously, these integrals are approximated using an MC estimate with 10,000 samples.

\paragraph{Time Prediction Results}

Results for the above mentioned next event time prediction task can be seen in Table~\ref{tab:time_pred}. While the results seem slightly in favor of the proposed MoE models compared to their decoder-only counterparts, they are by no means conclusive. What is important to note is that the results are, for the most part, similar between the two types of models. This indicates at the very least that the addition of personalization for a given neural MTPP will not harm its predictive power for timings of events (whereas it would consistently improve prediction performance of marks as seen previously). 

\begin{table}[]
    \centering
\caption{Mean next event prediction results for the predicted times $\hat t_{11}$ after conditioning on ten events in a given sequence. $L_1$ error is reported comparing true times to predicted times, with lower values being better. Superior performance between an MoE model and decoder-only counterpart is bolded for every dataset.}
\begin{tabular}{lcccc}
\hline
  & \multicolumn{4}{c}{Mean L1 Loss for Predicted Times ($\hat t_{11}$)}\\
Dataset & MoE-NHP & NHP & MoE-RMTPP & RMTPP\\
\hline\hline
Meme & \textbf{15.64} & 15.93 & \textbf{12.97} & 14.01 \\
Reddit & 4.57 & \textbf{4.51} & 3.88 & \textbf{3.87} \\
Amazon & \textbf{2.39} & 2.43 & \textbf{1.97} & 2.06 \\
LastFM & \textbf{0.56} & 0.61 & 0.33 & \textbf{0.32} \\
\hline
\end{tabular}
    \label{tab:time_pred}
\end{table}

\subsection{Sampling Experiments}

When modeling complex data with probabilistic models, having high log-likelihood scores does not always imply that the model will generate good samples \citep{theis2015note}. We therefore describe  experiments below that directly measure the sampling performance of the proposed models and baselines. 

\paragraph{Sampling Task} 
We sample future ``trajectories" (sequences of event times and marks) for different models, conditioned on a partial history of a sequence (of relative size $\rho, 0\% \le \rho \le 100\%$), and evaluate the quality of the sampled trajectories relative to the observed actual future trajectory for the same sequence.

More explicitly, for a given (real) test sequence $\mathcal{H}^s\in\mathcal{D}_{\text{Test}}$, let $\mathcal{H}^s_{\pi}$ be a portion of that sequence where $\pi < T$ is the smallest value that makes for $|\mathcal{H}^s_\pi| \approx \rho|\mathcal{H}^s|$ for $\rho \in [0,1]$. 
This partial, ground-truth sequence will be what we condition the model on and new events $(\hat{t},\hat{k}) $ will be sampled from time $\pi$ up until time $T$. We will denote $\mathcal{H}^s_{>\pi}$ as the portion of the real sequence not conditioned on, and $\hat{\mathcal{H}}^s_{>\pi}$ as the collection of all sampled events.

We use two different metrics to compare sampled data $\hat{\mathcal{H}}^s_{>\pi}$ to actual data $\mathcal{H}^s_{>\pi}$. These methods are introduced in the following two paragraphs.%

\paragraph{Sampled Marks Quality}

The first metric is a measurement of common marks between the two subsequences known as Jaccard Distance:
\begin{equation}
\text{JD}(\mathcal{H}, \hat{\mathcal{H}}) = 1-\frac{|\{k \in \mathcal{H}\}\cap\{\hat{k} \in \hat{\mathcal{H}}\}|}{|\{k \in \mathcal{H}\}\cup\{\hat{k} \in \hat{\mathcal{H}}\}|}.
\end{equation}
JD values close to 1  indicate that the  sampled out-of-distribution marks do not match well with the observed sequence (or the user that generated the sequence). Likewise, values close to 0 indicate appropriate marks being sampled. This  metric is particularly useful for datasets with a very large number of marks, with individual sequences only containing a fraction of them.

\paragraph{Sampled Times Quality}

The second metric is a measurement of how similar the empirical distributions of sampled timestamps are to the true timestamps. Here we use the Earth-movers (Wasserstein) distance,  defined for two sequences as:
\begin{equation}
\text{WD}(\mathcal{H}_{>\pi},\hat{\mathcal{H}}_{>\pi}) =\!\!\! \inf_{\gamma\in\Gamma(\{t\},\{\hat{t}\})} \int_{[\pi,T]\times[\pi,T]} \!|t-\hat{t}|d\gamma(t,\hat{t}),
\end{equation}
where $\{t\}$ and $\{\hat{t}\}$ are the empirical distributions of times in $\mathcal{H}_{>\pi}$ and $\hat{\mathcal{H}}_{>\pi}$ respectively, and $\Gamma$ is a set of joint probability distributions whose marginals are $\{t\}$ and $\{t'\}$.
WD values close to 0 indicate the two distributions are well-aligned  both in the timing of events and the number of them. Larger WD values indicate that the sampled times are less congruent with the original times for the given sequence, or more broadly, for the given user. 

\paragraph{Sampling Results}
We evaluated the two metrics averaged over 1000 randomly selected test sequences from all four datasets for $\rho$ values of 10\%, 30\%, and 50\%.

Table~\ref{tab:samples} shows the results. For JD, it appears that our personalization framework yields superior matching of the true mark mark distribution compared to the non-personalized, baseline models in 21 out of 24 settings. Similarly personalized models are only superior for mean WD values in 15 out of 24 settings. These findings further enforce our previous results for next event prediction, insofar as that the personalization framework appears to benefit mark distributions consistently versus yielding occasionally bettter modeling of the temporal dynamics. 

\begin{table}[]
    \centering
    \caption{(a) Mean Jacard distances and (b) mean Wasserstein distances for samples generated on the four datasets for varying percentages of sequences to condition ($\rho$) across the four model variants. Lower values are better for both, and bolbed values indicate superior performance between an MoE model and its decoder-only counterpart.}
    \resizebox{\columnwidth}{!}{
    \begin{tabular}{ccccccccccccc}
     \cline{1-6}\cline{8-13}
     \multicolumn{6}{c}{(a) Mean Sample Jacard Distances} & \hspace{-0.2em} & \multicolumn{6}{c}{(b) Mean Sample Wasserstein Distances} \\
     Dataset & $\rho$ & MoE-NHP & NHP & MoE-RMTPP & RMTPP & \hspace{-1em} & Dataset & $\rho$ & MoE-NHP & NHP & MoE-RMTPP & RMTPP \\
\cline{1-6}\cline{8-13} \multicolumn{13}{c}{} \\[-0.8\normalbaselineskip] \cline{1-6}\cline{8-13}
\vspace{-0.8em} \\
\multirow{3}{*}{Meme} & $10\%$ & \textbf{0.410} & 0.455 & 0.483 & \textbf{0.447} & \hspace{-1em} & \multirow{3}{*}{Meme} & $10\%$ & \textbf{26.302} & 27.133 & 28.296 & \textbf{28.022}\\
  & $30\%$ & \textbf{0.390} & 0.409 & 0.452 & \textbf{0.418} & \hspace{-1em} &   & $30\%$ & \textbf{21.477} & 22.862 & 23.768 & \textbf{23.484}\\
  & $50\%$ & \textbf{0.360} & 0.374 & 0.414 & \textbf{0.374} & \hspace{-1em} &   & $50\%$ & \textbf{17.657} & 18.204 & \textbf{18.547} & 18.842\\
\vspace{-0.5em}\\
\multirow{3}{*}{Reddit} & $10\%$ & \textbf{0.699} & 0.781 & \textbf{0.695} & 0.768 & \hspace{-1em} & \multirow{3}{*}{Reddit} & $10\%$ & 22.735 & \textbf{22.145} & \textbf{34.095} & 34.408\\
  & $30\%$ & \textbf{0.691} & 0.723 & \textbf{0.673} & 0.716 & \hspace{-1em} &   & $30\%$ & \textbf{18.765} & 18.826 & \textbf{25.019} & 25.113\\
  & $50\%$ & \textbf{0.650} & 0.677 & \textbf{0.656} & 0.684 & \hspace{-1em} &   & $50\%$ & 16.137 & \textbf{15.683} & 18.513 & \textbf{18.065}\\
\vspace{-0.5em}\\
\multirow{3}{*}{Amazon} & $10\%$ & \textbf{0.809} & 0.848 & \textbf{0.812} & 0.859 & \hspace{-1em} & \multirow{3}{*}{Amazon} & $10\%$ & \textbf{4.523} & 4.852 & \textbf{4.546} & 4.733\\
  & $30\%$ & \textbf{0.817} & 0.838 & \textbf{0.809} & 0.837 & \hspace{-1em} &   & $30\%$ & \textbf{3.718} & 4.020 & \textbf{3.979} & 4.046\\
  & $50\%$ & \textbf{0.830} & 0.836 & \textbf{0.827} & 0.841 & \hspace{-1em} &   & $50\%$ & \textbf{3.298} & 3.392 & \textbf{3.525} & 3.636\\
\vspace{-0.5em}\\
\multirow{3}{*}{LastFM} & $10\%$ & \textbf{0.534} & 0.556 & \textbf{0.503} & 0.530 & \hspace{-1em} & \multirow{3}{*}{LastFM} & $10\%$ & \textbf{3.755} & 3.822 & 3.402 & \textbf{3.351}\\
  & $30\%$ & \textbf{0.526} & 0.544 & \textbf{0.503} & 0.515 & \hspace{-1em} &   & $30\%$ & 3.132 & \textbf{3.015} & \textbf{2.793} & 2.843\\
  & $50\%$ & \textbf{0.552} & 0.556 & \textbf{0.540} & 0.544 & \hspace{-1em} &   & $50\%$ & 2.443 & \textbf{2.426} & 2.639 & \textbf{2.568}\\
\cline{1-6}\cline{8-13}
  \end{tabular}
    }
    \label{tab:samples}
\end{table}

\end{document}